# AI Risk Skepticism

**Roman V. Yampolskiy**
Computer Science and Engineering
University of Louisville
roman.yampolskiy@louisville.edu

**Abstract**
In this work, we survey skepticism regarding AI risk and show parallels with other types of scientific skepticism. We start by classifying different types of AI Risk skepticism and analyze their root causes. We conclude by suggesting some intervention approaches, which may be successful in reducing AI risk skepticism, at least amongst artificial intelligence researchers.

## 1. Introduction to AI Risk Skepticism

It has been predicted that if recent advancement in machine learning continue uninterrupted, human-level or even superintelligent Artificially Intelligent (AI) systems will be designed at some point in the near future [1]. Currently available (and near-term predicted) AI software is subhuman in its general intelligence capability but it is already capable of being hazardous in a number of narrow domains [2], mostly with regard to privacy, discrimination [3, 4], crime automation or armed conflict [5]. Superintelligent AI, predicted to be developed in the longer term, is widely anticipated [6] to be far more dangerous and is potentially capable of causing a lot of harm including an existential risk event for the humanity as a whole [7, 8]. Together the short-term and long-term concerns are known as AI Risk [9].

An infinite number of pathways exists to a state of the world in which a dangerous AI is unleashed [10]. Those include mistakes in design, programming, training, data, value alignment, self-improvement, environmental impact, safety mechanisms, and of course intentional design of Malevolent AI (MAI) [11-13]. In fact, MAI presents the strongest; some may say undeniable, argument against AI Risk skepticism (not to be confused with "skeptical superintelligence" [14]). While it may be possible to argue that a particular pathway to dangerous AI will not materialize or could be addressed, it seems nothing could be done against someone purposefully designing a dangerous AI. MAI convincingly establishes potential risks from intelligent software and makes denialist's point of view scientifically unsound. In fact, the point is so powerful, the authors of [11] were contacted by some senior researchers who expressed concern about impact such a publication may have on the future development and funding of AI research.

More generally, much can be inferred about the safety expectations for future intelligent systems from observing abysmal safety and security of modern software. Typically, users are required to click "Agree" on the software usage agreement, which denounces all responsibility from software developers and explicitly waves any guarantees regarding reliability and functionality of the provided software, including commercial products. Likewise, hardware components for the Internet of Things (IoT) notoriously lack security[1] in the design of the used protocol. Even in principle, sufficient levels of safety and security may not be obtainable for complex software products [15, 16].

Currently a broad consensus[2] exists in the AI Safety community, and beyond it, regarding importance of addressing existing and future AI Risks by devoting necessary resources to making AI safe and beneficial, not just capable. Such consensus is well demonstrated by a number of open letters[3,4,5,6] signed by thousands of leading practitioners and by formation of industry coalitions[7] with similar goals. Recognizing dangers posed by an unsafe AI a significant amount of research [17-24] is now geared to develop safety mechanisms for ever-improving intelligent software, with AI Safety research centers springing up at many top universities such as MIT[8], Berkeley[9], Oxford[10], and Cambridge[11], companies[12] and non-profits[13,14].

Given tremendous benefits associated with automation of physical and cognitive labor it is likely that funding and effort dedicated to creation of intelligent machines will only accelerate. However, it is important to not be blinded by the potential payoff, but to also consider associated costs. Unfortunately, like in many other domains of science, a vocal group of skeptics is unwilling to accept this inconvenient truth claiming that concerns about the human-caused issue of AI Risk are just "crypto-religious" [25] "pseudoscientific" "mendacious FUD" [26] "alarmism" [27] and "luddite" [28], "vacuous", "nonsense" [29], "fear of technology, opportunism, or ignorance", "anti-AI", "hype", "comical", "so ludicrous that it defies logic", "magical thinking", "techno-panic" [30], "doom-and-gloom", "Terminator-like fantasies" [31], "unrealistic", "sociotechnical blindness", "AI anxiety" [32], "technophobic", "paranoid" [33][15], "neo-fear" [34] and "mental masturbation" [35] by "fearmongers" [30], "AI Dystopians", "AI Apocalypsarians", - a "Frankenstein complex" [36]. They accuse AI Safety experts of being "crazy", "megalomaniacal", "alchemists", and "AI weenies", performing "parlor tricks" to spread their "quasi-sociopathic", "deplorable beliefs", about the "nerd Apocalypse", caused by their "phantasmagorical AI" [37].

Even those who have no intention to insult anyone have a hard time resisting such temptation: "The idea that a computer can have a level of imagination or wisdom or intuition greater than humans can only be imagined, in our opinion, by someone who is unable to understand the nature of human intelligence. It is not our intention to insult those that have embraced the notion of the technological singularity, but we believe that this fantasy is dangerous …" [38]. Currently, disagreement between AI Risk Skeptics [39] (interestingly Etzioni was an early AI Safety leader [40]) and AI Safety advocates [41] is limited to debate, but some have predicted that in the future it will become a central issue faced by humanity and that the so-called "species dominance debate" will result in a global war [42]. Such a war could be seen as an additional implicit risk from progress in AI.

---

[1] https://en.wikipedia.org/wiki/Internet_of_things#Security
[2] http://www.agreelist.org/s/advanced-artificial-intelligenc-4mtqyes0jrqy
[3] https://en.wikipedia.org/wiki/Open_Letter_on_Artificial_Intelligence
[4] https://futureoflife.org/open-letter-autonomous-weapons/
[5] https://futureoflife.org/ai-principles/
[6] https://futureoflife.org/ai-open-letter/
[7] https://www.partnershiponai.org
[8] https://futureoflife.org/
[9] https://humancompatible.ai/
[10] https://www.fhi.ox.ac.uk/
[11] https://www.cser.ac.uk/
[12] https://deepmind.com/
[13] https://openai.com/
[14] https://intelligence.org/
[15] Comment on article by Steven Pinker

"AI risk skeptics dismiss or bring into doubt scientific consensus of the AI Safety community on superintelligent AI risk, including the extent to which dangers are likely to materialize, severity of impact superintelligent AI might have on humanity and universe, or practicality of devoting resources to safety research." [43]. A more extreme faction, which could be called AI Risk Deniers[16], rejects any concern about AI-Risk, including from already deployed systems or soon to be deployed systems.

For example, 2009 AAAI presidential panel on Long-Term AI Futures tasked with review and response to concerns about the potential for loss of human control of computer-based intelligences, concluded: "The panel of experts was overall skeptical of the radical views expressed by futurists and science-fiction authors. Participants reviewed prior writings and thinking about the possibility of an "intelligence explosion" where computers one day begin designing computers that are more intelligent than themselves. They also reviewed efforts to develop principles for guiding the behavior of autonomous and semi-autonomous systems. Some of the prior and ongoing research on the latter can be viewed by people familiar with Isaac Asimov's Robot Series as formalization and study of behavioral controls akin to Asimov's Laws of Robotics. There was overall skepticism about the prospect of an intelligence explosion as well as of a "coming singularity," and also about the large-scale loss of control of intelligent systems." [44].

Denialism of anthropogenic climate change has caused dangerous delays in governments exercising control and counteraction. Similarly, influence of unchecked AI Risk denialism could be detrimental for the long term flourishing of human civilization as it questions importance of incorporating necessary safeguards into intelligent systems we are deploying. Misplaced skepticism has negative impact on allocating sufficient resources for assuring that developed intelligent systems are safe and secure. This is why it is important to explicitly call-out instances of AI risk denialism, just as it is necessary to fight denialism in other domains in which it is observed, such as history, healthcare and biology. In fact, in many ways the situation with advanced AI risk may be less forgiving. Climate change is comparable to soft takeoff [45], in which temperature is gradually rising by a few degrees over a 100-year period. An equivalent to superintelligence hard takeoff scenario would be global temperature rising by 100 degrees in a week.

## 2. Types of AI Risk Skeptics

---

[16] I first used the term "AI risk denier" in a 2015 paper https://arxiv.org/abs/1511.03246, and AGI risk skepticism in a 2014 (co-authored) paper: https://iopscience.iop.org/article/10.1088/0031-8949/90/1/018001/pdf

It is helpful to define a few terms and can be easily done by adopting the language addressing similar types of science denialism[17]: ***AI risk denial** is denial, dismissal, or unwarranted doubt that contradicts the scientific consensus on AI risk, including its effects on humanity. Many deniers self-label as "**AI risk skeptics**". AI risk denial is frequently implicit, when individuals or research groups accept the science but fail to come to terms with it or to translate their acceptance into action.* While denying risk from existing intelligent systems is pure denialism, with respect to future AIs, predicted to be superintelligent, it is reasonable to label such views as skepticism as evidence is not as strong for risk from such systems and as they don't currently exist and therefore are not subject to empirical testing for safety. Finally, we also introduce the concept of an "**AI safety skeptic"**, as someone who while accepting reality of AI risk doubts that safe AI is possible to achieve either in theory or at least in practice.

In order to overcome AI Risk Skepticism it is important to understand its causes and the culture supporting it. People who self-identify as AI Risk skeptics are very smart, ethical human beings and otherwise wonderful people, nothing in this paper should be interpreted as implying otherwise. Unfortunately, great people make great mistakes, and no mistake is greater than ignoring potential existential risk from development of advanced AI. In this section, I review most common reasons and beliefs for being an AI Risk Skeptic.

**Non-Experts** Non-AI-Safety researchers greatly enjoy commenting on all aspects of AI Safety. It seems like anyone who saw Terminator thinks they have sufficient expertise to participate in the discussion (on either side), but not surprisingly it is not the case. Not having formal training in the area of research should significantly discount importance of opinion of such public intellectuals, but it does not seem to be the case. By analogy, in discussions of cancer we listen to professional opinion of doctors trained in treating oncological diseases but feel perfectly fine ignoring opinions of business executives or lawyers. In AI Safety debates participants are perfectly happy to consider opinions of professional atheists [46], web-developers [37] or psychologists [47], to give just some examples.

**Wrong Experts** It may not be obvious but most expert AI Researchers are not AI Safety Researchers! Many AI Risk Skeptics are very knowledgeable and established AI researchers, but it is important to admit that having expertise in AI development is not the same as having expertise in AI Safety and Security. AI researchers are typically sub-domain experts in one of many sub-branches of AI research such as Knowledge Representation, Pattern Recognition, Computer Vision or Neural Networks, etc. Such domain expert knowledge does not immediately make them experts in all other areas of artificial intelligence, AI Safety being no exception. More generally, a software developer is not necessarily a cybersecurity expert. It is easy to illustrate this by analogy with a non-computer domain. For example, a person who is an expert on all things related to cement is not inevitable an expert on the placement of emergency exits even though both domains have a lot to do with building construction.

**Professional Skeptics** Members of skeptic organizations are professionally predisposed to question everything and it is not surprising that they find claims about properties of future superintelligent machines to fall in their domain of expertise. For example, Michael Shermer, founder of Skeptics Society and publisher of Skeptic magazine, has stated [46]: "I'm skeptical. … all such doomsday scenarios involve a long sequence of if-then contingencies, a failure of which at any point would negate the apocalypse." Similarly, those who are already skeptical about one domain of science, for example theory of evolution, are more likely to also exhibit skepticism about AI Risk[18].

---

[17] https://en.wikipedia.org/wiki/Climate_change_denial

**Ignorant of Literature** Regardless of background, intelligence or education many commentators on AI Safety seem to be completely unaware of literature on AI Risk, top researchers in the field, their arguments and concerns. It may no longer be possible to read everything on the topic due to the sheer number of publications produced in recent years, but there really is no excuse for not familiarizing yourself with top books [7, 8, 48] or survey papers [9] on the topic. It is of course impossible to make a meaningful contribution to the discussion if one is not aware of what is actually being discussed or is engaging with the strawman arguments.

**Skeptics of Strawman**
Some AI skeptics may actually be aware of certain AI Safety literature, but because of poor understanding or because they were only exposed to weaker arguments for AI Risk, they find them unconvincing or easily dismissible and so feel strongly justified in their skeptical positions. Alternatively, they may find a weakness in one particular pathway to dangerous AI and consequently argue against "fearing the reaper" [49].

**With Conflict of Interest and Bias** Lastly, we cannot ignore an obvious conflict of interest many AI researchers, tech CEOs, corporations and others in the industry have with regards to their livelihood and the threat AI Risk presents to unregulated development of intelligent machines. History teaches us that we can't count on people in the industry to support additional regulation, reviews or limitations against their direct personal benefit. Tobacco company representatives, for years, assured the public that cigarettes are safe, non-carcinogenic and non-addictive. Oil companies rejected any concerns public had about connection between burning of fossil fuels and global climate change, despite knowing better.

It is very difficult for a person whose success, career, reputation, funding, prestige, financial well-being, stock options and future opportunities depend on unobstructed development of AI to accept that the product they are helping to develop is possibly unsafe, requires government regulation, and internal or even external review boards. As Upton Sinclair put it: "It is difficult to get a man to understand something when his salary depends upon his not understanding it." They reasonably fear that any initial concessions may lead to a significant "safety overhead" [50], reduced competitiveness, slowdown in progress or a moratorium on development (à la human cloning) or even an outright ban on future research. The conflict of interest developers of AI have with respect to their ability to impartially assess dangers of their products/services is unquestionable and would be flagged by any ethics panel. Motivated misinformation targeting lay people, politicians, and public intellectuals may also come from governments, thought leaders and activist citizens interested in steering debate in particular directions [51]. Corporations may additionally worry about legal liability and overall loss of profits.

In addition to the obvious conflicts of interest, most people, including AI researchers, are also subject to a number of cognitive biases making them underappreciate AI Risk. Those would include Optimism Bias (thinking that you are at a smaller risk of suffering a negative outcome), and Confirmation Bias (interpreting information in a way that confirms preconceptions). Additionally, motivated reasoning may come into play, as Baum puts it [51]: "Essentially, with their sense of self-worth firmed up, they become more receptive to information that would otherwise threaten their self-worth. As a technology that could outperform humans, superintelligence could pose an especially pronounced threat to people's sense of self-worth. It may be difficult for people to feel good and efficacious if they would soon be superseded by computers. For at least some people, this could be a significant reason to reject information about the prospect of superintelligence, even if that information is true."

---

[18] https://web.archive.org/web/20120611073509/http://www.discoverynews.org/2011/02/artificial_intelligence_is_not044151.php

## 3. Arguments for AI Risk Skepticism

In this section, we review most common arguments for AI risk skepticism. Russell has published a similar list, which in addition to objections to AI risk concerns also includes examples of flawed suggestions for assuring AI safety [52], such as: "Instead of putting objectives into the AI system, just let it choose its own", "Don't worry, we'll just have collaborative human-AI teams", "Can't we just put it in a box?", Can't we just merge with the machines?" and "Just don't put in 'human' goals like self-preservation".

Importance of understanding denialists' mindset is well-articulated by Russell: "When one first introduces [AI risk] to a technical audience, one can see the thought bubbles popping out of their heads, beginning with the words "But, but, but . . ." and ending with exclamation marks. The first kind of *but* takes the form of denial. The deniers say, "But this can't be a real problem, because XYZ." Some of the XYZs reflect a reasoning process that might charitably be described as wishful thinking, while others are more substantial. The second kind of *but* takes the form of deflection: accepting that the problems are real but arguing that we shouldn't try to solve them, either because they're unsolvable or because there are more important things to focus on than the end of civilization or because it's best not to mention them at all. The third kind of *but* takes the form of an oversimplified, instant solution: "But can't we just do ABC?" As with denial, some of the ABCs are instantly regrettable. Others, perhaps by accident, come closer to identifying the true nature of the problem. … Since the issue seems to be so important, it deserves a public debate of the highest quality. So, in the interests of having that debate, and in the hope that the reader will contribute to it, let me provide a quick tour of the highlights so far, such as they are." [53].

In addition to providing a comprehensive list of arguments for AI risk skepticism, we have also classified such objections into six categories (see Figure 1): Objections related to Priorities, Technical issues, AI Safety, Ethics, Bias, and Miscellaneous ones. While research on types of general skepticism exists [54], to the best of our knowledge this is the first such taxonomy specifically for AI risk. In general, we can talk about politicized skepticism and intellectual skepticism [55]. Politicized skepticism has motives other than greater understanding, while intellectual skepticism aims for better comprehension and truth seeking. Our survey builds and greatly expands on previous lists from Turing [56], Baum [55], Russell [52], and Ceglowski [37].

| **PRIORITIES OBJECTIONS** |
|---|
| <ul><li>Too Far</li><li>Soft Takeoff is more likely and so we will have Time to Prepare</li><li>No Obvious Path to Get to AGI from Current AI</li><li>Short Term AI Concerns over AGI Safety</li><li>Something Else is More Important</li></ul> |
| **TECHNICAL OBJECTIONS** |
| <ul><li>AI Doesn't Exist</li><li>Superintelligence is Impossible</li><li>Self-Improvement is Impossible</li><li>AI Can't be Conscious</li><li>AI Can be Just a Tool</li><li>We can Always just Turn it Off</li><li>We Can Reprogram AIs if We Don't Like What They Do</li><li>AI doesn't have a Body and so can't Hurt Us</li><li>If AI is as Capable as You Say, it Will not Make Dumb Mistakes</li><li>Superintelligence Would (Probably) Not Be Catastrophic</li><li>Self-preservation and Control Drives Don't Just Appear They Have to be Programmed In</li><li>An AI is not Pulled at Random from the Mind Design Space</li><li>AI Can't Generate Novel Plans</li></ul> |
| **AI SAFETY OBJECTIONS** |
| <ul><li>AI Safety Can't be Done Today</li><li>AI Can't be Safe</li><li>Skepticism of Particular Risks</li><li>Skepticism of Particular Safety Methods</li><li>Skepticism of Researching Impossibility Results</li></ul> |
| **ETHICAL OBJECTIONS** |
| <ul><li>Superintelligence is Benevolence</li><li>Let the Smarter Beings Win</li><li>Let's Gamble</li><li>Malevolent AI is not worse than Malevolent Humans</li></ul> |
| **BIASED OBJECTIONS** |
| <ul><li>AI Safety Researchers are Non-Coders</li><li>Majority of AI Researchers is not Worried</li><li>Anti-Media Bias</li><li>Keep it Quiet</li><li>Safety Work just Creates an Overhead Slowing Down Research</li><li>Heads in the Sand</li></ul> |
| **MISCELLANEOUS OBJECTIONS** |
| <ul><li>So Easy it will be Solved Automatically</li><li>AI Regulation Will Prevent Problems</li><li>Other Arguments, …</li></ul> |

Figure 1: Taxonomy of Objections to AI Risk

## 3.1 Priorities Objections

**Too Far** A frequent argument against work on AI Safety is that we are hundreds if not thousands of years away from developing superintelligent machines and so even if they may present some danger it is a waste of human and computational resources to allocate any effort to address Superintelligence Risk at this point in time. Such position doesn't take into account possibility that it may take even longer to develop appropriate AI Safety mechanisms and so the perceived abundance of time is a feature, not a bug. It also ignores a non-zero possibility of an earlier development of superintelligence.

**Soft Takeoff is more likely and so we will have Time to Prepare** AI takeoff refers to the speed with which an AGI can get to superintelligent capabilities. While hard takeoff is likely and means that process will be very quick, some argue that we will face a soft takeoff and so will have adequate time (years) to prepare [45]. While nobody knows the actual take off speed at this point, it is prudent to be ready for the worst-case scenario.

**No Obvious Path to Get to AGI from Current AI** While we are making good progress on AI, it is not obvious how to get from our current state in AI to AGI and current methods may not scale [57]. This may be true, but this is similar to the "Too Far" objection and we definitely need all the time possible to develop necessary safety mechanisms. Additionally, current state-of-the-art systems [58], don't seem to hit limits yet, subject to availability of compute for increasing model size [59, 60].

**Something Else is More Important** Some have argued that global climate change, pandemics, social injustice, and a dozen of other more immediate concerns are more important than AI risk and should be prioritized over wasting money and human capital on something like AI Safety. But, development of safe and secure superintelligence is a possible meta-solution to all the other existential threats and so resources allocated to AI risk are indirectly helping us address all the other important problems. Time wise it is also likely, that AGI will be developed before projected severe impact from such issues as global climate change.

**Short Term AI Concerns over AGI Safety** Similar to the argument that something else is more important, proponents claim that immediate issues with today's AIs, such as algorithmic bias, technological unemployment or limited transparency should take precedence over concerns about future technology (AGI/superintelligence), which doesn't yet exist and may not exist for decades [61].

## 3.2 Technical Objections

**AI Doesn't Exist** The argument is that current developments in Machine Learning are not progress in AI, but are just developments in statistics, particularly in matrix multiplication and gradient descent[19]. Consequently, it is suggested that calls for regulation of AI are absurd. Of course, human criminal behavior can be seen as interactions of neurotransmitters and ion channels, making their criminalization questionable.

---
[19] https://twitter.com/benhamner/status/892136662171504640

**Superintelligence is Impossible** If a person doesn't think that superintelligence can ever be built they will of course view Risk from Superintelligence with strong skepticism. Most people in this camp assign a very small (but usually not zero) probability to the actual possibility of superintelligent AI coming into existence [62-64], but if even a tiniest probability is multiplied by the infinite value of the Universe the math seems to be against skepticism. Skeptics in this group will typically agree that if superintelligence did exist it would have potential of being harmful. "Within the AI community, a kind of denialism is emerging, even going as far as denying the possibility of success in achieving the long-term goals of AI. It's as if a bus driver, with all of humanity as passengers, said, "Yes, I am driving as hard as I can towards a cliff, but trust me, we'll run out of gas before we get there!"" [53].

**Self-Improvement is Impossible** This type of skepticism concentrates on the supposed impossibility of intelligence explosion, as a side-effect of recursive self-improvement [65], due to fundamental computational limits [49] and software complexity [66]. Of course such limits are not a problem as long as they are actually located above the level of human capabilities.

**AI Can't be Conscious** Proponents argue that in order to be dangerous AI has to be conscious [67]. As AI risk is not predicated on artificially intelligent systems experiencing qualia [68, 69], it is not relevant if the system is conscious or not. This objection is as old as the field of AI itself, as Turing addressed "The Argument from Consciousness" in his seminal paper [56].

**AI Can be Just a Tool** A claim that we do not need A(General)I to be an independent agent, it is sufficient for them to be designed as assistants to humans in particular domains, such as GPS navigation and so permit as to avoid dangers of fully independent AI[20]. It is easy to see that the demarcation between Tool AI and AGI is very fuzzy and likely to gradually shift as capability of the tool increases and it obtains additional capabilities[21].

**We can Always just Turn it Off** A very common argument of AI risk skeptics is that any misbehaving AI can be simply turned off, so we have nothing to worry about [70]. If skeptics realize that modern computer viruses are a subset of very low capability malevolent AIs it becomes obvious why saying "just turn it off" may not be a practical solution.

**We Can Reprogram AIs if We Don't Like What They Do** Similar to the idea of turning AI off, is the idea that we can reprogram AIs if we are not satisfied with their performance [71]. Such "in production" correction is equally hard to accomplish as it can be shown to be equivalent to shutting current AI off.

**AI doesn't have a Body and so can't Hurt Us** This is a common argument and it completely ignores the realities of modern ultra-connected world. Given simple access to the internet, it is easy to affect the world via hired help, digital currencies, internet of things, cyberinfrastructure or even DNA synthesis [72].

---

[20] https://wiki.lesswrong.com/wiki/Tool_AI
[21] http://lesswrong.com/lw/cze/reply_to_holden_on_tool_ai/

**If AI is as Capable as You Say, it Will not Make Dumb Mistakes** How can superintelligence not understand what we really want? This seems like a paradox [21], any system worthy of the title "human-level" must have the same common sense as we do [73]. Unfortunately, an AI could be a very powerful optimizer while at the same time not being aligned with goals of humanity [74, 75].

**Superintelligence Would (Probably) Not Be Catastrophic** Not quite benevolent, but superintelligence would not be very dangerous by default, or at least the dangers would not be catastrophic [76] or its behavior would be correctable in time and is unlikely to be malevolent if not explicitly programmed to be [77]. Some of the ideas in [76] are analyzed in the highly relevant paper on modeling and interpreting expert disagreement about AI [78].

**Self-preservation and Control Drives Don't Just Appear They Have to be Programmed In** LeCun has publicly argued that "the desire to control access to resources and to influence others are drives that have been built into us by evolution for our survival. There is no reason to build these drives into our AI systems. Some have said that such drives will spontaneously appear as sub-goals of whatever objective we give to our AIs. Tell a robot "get me coffee" and it will destroy everything on its path to get you coffee, perhaps figuring out in the process how to prevent every other human from turning it off. We would have to simultaneously be extremely talented engineers to build such an effective goal-oriented robot, and extremely stupid and careless engineers to not put any obvious safeguards into its objective to ensure that it behaves properly."[22] This dismisses research which indicates that such AI drives do appear due to game theoretic and economic reasons [79].

**An AI is not Pulled at Random from the Mind Design Space** Kruel has previously argued that "[a]n AI is the result of a research and development process. A new generation of AI's needs to be better than other products at *"Understand What Humans Mean"* and *"Do What Humans Mean"* in order to survive the research phase and subsequent market pressure." [80]. Of course, being better doesn't mean being perfect or even great, almost all existing software is evidence of very poor quality of the software research/development process.

**AI Can't Generate Novel Plans** As originally stated by Ada Lovelace: "The Analytical Engine has no pretensions whatever to originate anything. It can do whatever we know how to order it to perform. It can follow analysis; but it has no power of anticipating any analytical relations or truths. Its province is to assist us to making available what we are already acquainted with." [81]. Of course, numerous counterexamples from modern AI [82] systems provide a counterargument by existence. This doesn't stop modern scholars from making similar claims, specifically arguing that only humans can have "curiosity, imagination, intuition, emotions, passion, desires, pleasure, aesthetics, joy, purpose, objectives, goals, telos, values, morality, experience, wisdom, judgment, and even humor." [38]. Regardless of ongoing work [83], most AI safety researchers are not worried about deadly superintelligence not having a superior sense of humor.

## 3.3 AI Safety Related Objections

---

[22] https://www.facebook.com/yann.lecun/posts/10154220941542143

**AI Safety Can't be Done Today** Some people may agree with concerns about superintelligence but argue that AI Safety work is not possible in the absence of a superintelligent AI on which to run experiments [37]. This view is contradicted by a significant number of publications produced by the AI Safety community in recent years, and the author of this article (and his co-authors) in particular [8, 19, 84-89].

**AI Can't be Safe** Another objection to doing AI Safety work is based on publications showing that fundamental aspects of the control problem [90], such as containment [91], verification [16], or morality [92] are simply impossible to solve and so such research is a wasted effort. Solvability of the control problem in itself is one of the most important open questions in AI Safety, but not trying is the first step towards failure.

**Skepticism of Particular Risks** Even people troubled by some AI Risks may disagree about specific risks they are concerned about and may disagree on safety methods to implement, which ones are most likely to be beneficial and which ones are least likely to have undesirable side effects. This is something only additional research can help resolve.

**Skepticism of Particular Safety Methods** AI companies may be dismissive of effectiveness of risk mitigation technology developed by their competitors in the hopes of promoting and standardizing their own technology [55]. Such motivated skepticism should be dismissed.

**Skepticism of Researching Impossibility Results** Doing work on theoretical impossibility results [93-95] in AI safety may not translate to problems in practice, or to at least not be as severe as predicted. However, such research may cause reductions in funding for safety work or to cause new researchers to stay away from the field of AI safety, but this is not an argument against importance of AI risk research in general.

### 3.4 Ethical Objections

**Superintelligence is Benevolence** Scholars observed that as humans became more advanced culturally and intellectually they also became nicer, less violent, and more inclusive [47]. Some have attempted to extrapolate from that pattern to superintelligent advanced AIs that they will also be benevolent to us and our habitat [96] and will not develop their own goals which are not programmed into them explicitly. However, superintelligence doesn't imply benevolence [97], which is directly demonstrated by the Bostrom's Orthogonality Thesis [74, 75].

**Let the Smarter Beings Win** This type of skeptic doesn't deny that superintelligent system will present a lot of risk to humanity but argue that if humanity is replaced with a more advanced sentient beings it will be an overall good thing. They give very little value to humanity and see people as mostly having a negative impact on the planet and cognition in the universe. Similarly, AI rights advocates argue that we should not foist our values on our mind children because it would be a type of forced assimilation. Majority of AI researchers don't realize that people with such views are real, but they are and some are also AI researchers. For example, de Garis [42] has argued that humanity should make room for superintelligent beings. Majority of humanity is not on board with such self-destructive outcomes, perhaps because of a strong inherent pro-human bias.

**Let's Gamble** In the vast space of possible intelligences [98] some are benevolent, some are neutral and others are malicious. It has been suggested that only a small subset of AIs are strictly malevolent and so we may get lucky and produce a neutral or beneficial superintelligence by pure chance. Gambling with future of human civilization doesn't seem like a good proposition.

**Malevolent AI is not worse than Malevolent Humans** The argument is that it doesn't matter who is behind malevolent action, human actors or AI, the impact is the same [33]. Of course, a more intelligent and so more capable AI can be much more harmful and is harder to defeat with human resources, which are frequently sufficient to counteract human adversaries. AI is also likely to be cognitively different from humans and so find surprising ways to cause harm.

### 3.5 Biased Objections

**AI Safety Researchers are Non-Coders** An argument is frequently made that since many top AI Safety researchers do not write code, they are unqualified to judge AI Risk or its correlates[23]. However, one doesn't need to write code in order to understand the inherent risk of AGI, just like someone doesn't have to work in a wet lab to understand dangers of pandemics from biological weapons.

**Majority of AI Researchers is not Worried** To quote from Dubhashi and Lappin - "While it is difficult to compute a meaningful estimate of the probability of the singularity, the arguments here suggest to us that it is exceedingly small, at least within the foreseeable future, and this is the view of most researchers at the forefront of AI research." [99]. Not only does this misrepresent actual views of actual AI researchers [100, 101], it is also irrelevant, even if 100% of mathematicians believed $2 + 2 = 5$, it would still be wrong. Scientific facts are not determined by democratic process, and you don't get to vote on reality or truth.

**Anti-Media Bias** Because of how the media sensationalizes coverage of AI Safety issues, it is also likely that many AI Researchers have Terminator-aversion, subconsciously or explicitly equating all mentions of AI Risk with pseudoscientific ideas from Hollywood blockbusters. While, literal "Terminators" are of little concern to the AI safety community, AI weaponized for military purposes is a serious challenge to human safety.

**Keep it Quiet** It has been suggested that bringing up concerns about AI risk may jeopardize AI research funding and bring on government regulation. Proponents argue that it is better to avoid public discussions of AI risk and capabilities, which advanced AI may bring, as it has potential of bringing on another AI "winter". There is also some general concern about the reputation of the field of AI [55].

**Safety Work just Creates an Overhead Slowing Down Research** Some developers are concerned that integrating AI safety into research will create a significant overhead and make their projects less competitive. The worry is that groups which don't worry about AI risk will get to human-level AI faster and cheaper. This is similar to cost cutting measure in software development, where security concerns are sacrificed, to be the first to the market.

---
[23] http://reducing-suffering.org/predictions-agi-takeoff-speed-vs-years-worked-commercial-software

**Heads in the Sand** An objection from Turing's classic paper [56] arguing that "The consequences of machines thinking would be too dreadful. Let us hope and believe that they cannot do so." And his succinct response "I do not think that this argument is sufficiently substantial to require refutation." [56]. In the same paper Turing describes and appropriately dismisses a number of common objections to the possibility of machines achieving human level performance in thinking: The Theological Objection, The Mathematical Objection, the Argument from Various Disabilities, Lady Lovelace's Objection, Argument from Continuity of the Nervous System, The Argument from Informality of Behavior, and even the Argument from Extrasensory Perception [56].

### 3.6 Miscellaneous Objections

**So Easy it will be Solved Automatically** Some scholars think that the AI risk problem is trivial and will be implicitly solved as a byproduct of doing regular AI research [102]. Same flawed logic can be applied to other problems such as cybersecurity, but of course, they never get completely solved, even with significant effort.

**AI Regulation Will Prevent Problems** The idea is that we don't need to worry about AI Safety because government regulation will intervene and prevent problems. Given how poorly legislation against hacking, computer viruses or even spam has performed it seems unreasonable to rely on such measures for prevention of AI risk.

**Other Arguments**
There are many other arguments by AI risk skeptics, which are so weak they are not worth describing, but the names of arguments hint at their quality, for example: The arguments from Wooly Definitions, Einstein's Cat, Emus, Slavic Pessimism, My Roommate, Gilligan's Island, Transhuman Woodoo, and Comic Books [37]. Luckily, others have taken the time to address them [103, 104] so we did not have to.

Russell provides examples of what he calls "Instantly regrettable remarks", statements from AI researchers which they are likely to retract after some retrospection [53]. He follows each one with a refutation, but that seems unnecessary given low quality of the original statements:

- "Electronic calculators are superhuman at arithmetic. Calculators didn't take over the world; therefore, there is no reason to worry about superhuman AI."
- "Horses have superhuman strength, and we don't worry about proving that horses are safe; so we needn't worry about proving that AI systems are safe."
- "Historically, there are zero examples of machines killing millions of humans, so, by induction, it cannot happen in the future."
- "No physical quantity in the universe can be infinite, and that includes intelligence, so concerns about superintelligence are overblown."
- "We don't worry about species-ending but highly unlikely possibilities such as black holes materializing in near-Earth orbit, so why worry about superintelligent AI?"

While aiming for good coverage of the topic of AI risk skepticism we have purposefully stopped short of analyzing every variant of the described main types of arguments as the number of such objections continues to grow exponentially and it is not feasible or even desirable to include everything into a survey. Readers who want to get deeper into the debate may enjoy the following articles [105-113]/videos [114, 115]. In our future work we may provide additional analysis of the following objections:

- Bringing up concerns about AGI may actively contribute to the public misunderstanding of science and by doing so contribute to general science denialism.
- Strawman objections: "The thought that these systems would wake up and take over the world is ludicrous." [29].
- We will never willingly surrender control to machines.
- While AGI is likely, superintelligence is not.
- Risks from AI are minuscule in comparison to benefits (immortality, free labor, etc.) and so can be ignored.
- "Intelligence is not a single dimension, so "smarter than humans" is a meaningless concept." [116].
- "Humans do not have general purpose minds, and neither will AIs." [116].
- "Emulation of human thinking in other media will be constrained by cost." [116].
- "Dimensions of intelligence are not infinite." [116].
- "Intelligences are only one factor in progress." [116].
- You can't control research or ban AI [53].
- Malevolent use of AI is a human problem, not a computer problem [46].
- "Speed alone does not bring increased intelligence" [117].
- Not even exponential growth of computational power can reach the level of superintelligence [118].
- AI risk researchers are uneducated/conspiracy theorists/crazy/etc, so they are wrong.
- AI has been around for 65 years and didn't destroy humanity, it is unlikely to do so in the future.
- AI risk is science fiction.
- Just box it; just give it laws to follow; just raise it as a human baby; just …
- AI is just a tool, it can't generate its own goals because it is not conscious.
- I don't want to make important AI researchers angry at me and retaliate against me.
- Narrow AI/robots can't even do some basic thing, certainly they can't present danger to humanity.
- Real threat is AI being too dumb and making mistakes.
- Certainly, many smart people are already working on AI safety they will take care of it.
- Big companies like Google or Microsoft would never release a dangerous product or service which may damage their reputation or reduce profits.
- Smartest person in the world, [multiple names are used by proponents], is not worried about it so it must not be a real problem.

## 4. Countermeasures for AI Risk Skepticism

First, it is important to emphasize that just like with any other product or service the burden of proof [119] is on the developers/manufacturers (frequently AI Risk skeptics) to show that their AI will be safe and secure regardless of its capability, customization, learning, domain of utilization or duration of use. Proving that an intelligent agent in a novel environment will behave in a particular way is a very high standard to meet. The problem could be reduced to showing that a particular human or an animal, for example a Pit bull, is safe to everyone, a task long known to be impractical. It seems to be even harder with much more capable agents, such as AGI. The best we can hope for is showing some non-zero probability of safe behavior.

A capable AI researcher not concerned with safety is very dangerous. It seems that the only solution to reduce prevalence of AI risk denialism is education. It is difficult for a sharp mind to study the best AI risk literature and to remain unconvinced of scientific merits behind it. The legitimacy of risk from uncontrolled AI is undeniable. This is not fear mongering, we don't have an adequate amount of fear in the AI researcher community, an amount which would be necessary to make sure that sufficient precautions are taken by everyone involved. Education is likewise suggested as a desirable path forward by the skeptics, so all sides agree on importance of education. Perhaps if we were to update and de-bias recommendations from the 2009 AAAI presidential panel on Long-Term AI Futures to look like this: "The group suggested outreach and communication to people and organizations about the ~~low~~ likelihood of the radical outcomes, sharing the rationale for the overall ~~comfort~~ [position] of scientists in this realm, and for the need to educate people outside the AI research community about ~~the promise of~~ AI" [44], we could make some progress on AI risk denialism reduction.

The survival of humanity could depend on rejecting superintelligence misinformation [51]. Two main strategies could be identified: those aimed at preventing spread of misinformation and those designed to correct peoples' understanding after exposure to misinformation. Baum reviews some ways to prevent superintelligence misinformation, which would also apply to reducing AI Risk skepticism [51]: educate prominent voices, create reputation costs, mobilize against institutional misinformation, focus media attention on constructive debate, establish legal requirements. For correcting superintelligence misinformation Baum suggests: building expert consensus and the perception of thereof, address pre-existing motivations for believing misinformation, inoculate with advance warnings, avoid close association with polarizing ideas, explain misinformation and corrections [51].

Specifically for politicized superintelligence skepticism Baum suggests [55]: "With this in mind, one basic opportunity is to raise awareness about politicized skepticism within communities that discuss superintelligence. Superintelligence skeptics who are motivated by honest intellectual norms may not wish for their skepticism to be used politically. They can likewise be cautious about how to engage with potential political skeptics, such as by avoiding certain speaking opportunities in which their remarks would be used as a political tool instead of as a constructive intellectual contribution. Additionally, all people involved in superintelligence debates can insist on basic intellectual standards, above all by putting analysis before conclusions and not the other way around. These are the sorts of things that an awareness of politicized skepticism can help with." Baum also recommends [55] to: "redouble efforts to build scientific consensus on superintelligence, and then to draw attention to it", "engage with AI corporations to encourage them to avoid politicizing skepticism about superintelligence or other forms of AI", and "follow best practices in debunking misinformation in the event that superintelligence skepticism is politicized." "… Finally, the entire AI community should insist that policy be made based on an honest and balanced read of the current state of knowledge. Burden of proof requirements should not be abused for private gain. As with climate change and other global risks, the world cannot afford to prove that superintelligence would be catastrophic. By the time uncertainty is eliminated, it could be too late." [55].

AI risk education research [120] indicates that most AI risk communication strategies are effective [121] and are not counter-productive and the following "good practices" work well for introducing general audiences to AI risk [120]: "

1. Allow the audience to engage in guided thinking on the subject ("What do you think the effects of human-level AI will be?"), but do not neglect to emphasize its technical nature
2. Reference credible individuals who have spoken about AI risk (such as Stephen Hawking, Stuart Russell, and Bill Gates)
3. Reference other cases of technological risk and revolution (such as nuclear energy and the Industrial Revolution)
4. Do not reference science-fiction stories, unless, in context, you expect an increase in the audience's level of engagement to outweigh a drop in their perceptions of the field's importance and its researchers' credibility
5. Do not present overly vivid or grave disaster scenarios
6. Do not limit the discussion to abstractions (such as "optimization," "social structures," and "human flourishing"), although they may be useful for creating impressions of credibility"

Recent research indicates that individual differences in the AI risk perception may be personality [122] and/or attitude [123, 124] dependent but are subject to influence by experts [125] and choice of language [126].

Healthy skepticism is important to keep scientists, including AI researchers honest. For example, during early days of AI research it was predicted that human level performance will be quickly achieved [127]. Luckily, a number of skeptics [128, 129] argued that perhaps the problem is not as simple as it seems, bringing some conservativism to the overly optimistic predictions of researchers and as a result improving quality of research actually being funded and conducted by AI researchers. For a general overview of threat inflation Thierer's work on technopanics [130] is a good reference.

## 5. Conclusions

In this paper, we didn't reiterate most of the overwhelming evidence for AI Risk concerns, as it was outside of our goal of analyzing AI Risk skepticism. Likewise, we did not go in depth with rebuttals to every type of objections to AI Risk. It is precisely because of skeptic attitudes from the majority of mainstream AI researchers that the field of AI Safety was born outside of academia [131]. Regardless, AI Risk skeptics need to realize that the burden of proof is not on AI Safety researchers to show that technology may be dangerous but on AI developers to establish that their technology is safe at the time of deployment and throughout its lifetime of operation. Furthermore, while science operates as a democracy (via majority of peer-reviewers), the facts are not subject to a vote. Even if AI Safety researchers comprise only a small minority of the total number of AI researchers that says nothing about the true potential of intelligent systems for harmful actions. History is full of examples (continental drift [132], quantum mechanics [133]) in which a majority of scientists held a wrong view right before a paradigm shift in thinking took place. Since, just like AI Skeptics, AI Safety researchers also have certain biases, to avoid pro or con prejudice in judgment it may be a good idea to rely on impartial juries of non-peers (scientists from outside the domain) whose only job would be to evaluate evidence for a particular claim.

It is obvious that designing a Safe AI is a much harder problem than designing an AI and so will take more time. The actual time to human level AI is irrelevant; it will always take longer to make such an AI human friendly. To move the Overton window on AI Risk, AI Safety researchers have to be non-compromising in their position. Perhaps a temporary moratorium on AGI (but not AI) research similar to the one in place for human cloning needs to be considered. It would boost our ability to engage in differential technological development [134-136] increasing our chances of making the AGI safe. AI Safety research definitely needs to get elevated priority and more resources including funding and human capital. Perhaps AI Safety researchers could generate funding via economic incentives from developing safer products. It may be possible to market "Safe AI Inside" government certification on selected progressively ever-smarter devices to boost consumer confidence and sales. This would probably require setting up "FDA for algorithms" [137, 138].

Scientific skepticism in general and skepticism about predicted future events is of course intellectually defensible and is frequently desirable to protect again flawed theories [130]. However, it is important to realize that 100% proof is unlikely to be obtained in some domains and so a Precautionary Principle (PP) [139] should be used to protect humanity against existential risks. Holm and Harris, in their skeptical paper, define PP as follows [140]: "When an activity raises threats of serious or irreversible harm to human health or the environment, precautionary measures that prevent the possibility of harm shall be taken even if the causal link between the activity and the possible harm has not been proven or the causal link is weak and the harm is unlikely to occur." To use a stock market metaphor, no matter how great a return on investment one is promised, one should not ignore the possibility of losing the principal.

**Acknowledgements**

The author is grateful to Seth Baum for sharing a lot of relevant literature and providing feedback on an early draft of this paper. In addition, author would like to acknowledge his own bias, as an AI safety researcher I would benefit from flourishing of the field of AI safety. I also have a conflict of interest, as a human being with a survival instinct I would benefit from not being exterminated by uncontrolled AI.